\documentclass[10pt, a4paper]{llncs} 
\usepackage[latin1]{inputenc}
\usepackage[english]{babel}
\usepackage{epic}
\usepackage{epstopdf}
\usepackage{graphicx}
\usepackage{url}
\usepackage{mathptmx}
\usepackage{palatino}
\usepackage{indentfirst}
\usepackage{amssymb}
\usepackage{amsmath}
\usepackage{textcomp}
\usepackage{natbib}
\usepackage{gb4e} 

\newcommand\Land{\&^\Pi}

\newcommand\et\land
\newcommand\fl{\rightarrow}

\newcommand\ma[1]{"\emph{#1}"}

\newcommand\ttt{\mathbf{t}}
\newcommand\eee{\mathbf{e}}
\newcommand\sss{\mathbf{s}}

\newcommand\event{\mathbf{v}}

\begin{document}
\title{Semantic Types, Lexical Sorts and Classifiers}
\author{Bruno Mery \& Christian Retoré} 
\institute{Université de Bordeaux, IRIT-CNRS, LABRI-CNRS\thanks{This research has benefitted from grants and inputs by ANR Polynomie, Project Itipy (Région Aquitaine), and CoLAn. We are indebted to the reviewers for their input.}}

\maketitle

\begin{abstract}
We propose a cognitively and linguistically motivated set of sorts for lexical semantics in a compositional setting: the classifiers in languages that do have such pronouns. These sorts are needed to include lexical considerations in a semantical analyser such as Boxer or Grail. 
Indeed, all proposed lexical extensions of usual Montague semantics to model restriction of selection, felicitous and infelicitous copredication require a rich and refined type system whose base types are the lexical sorts, the basis of the many-sorted logic in which semantical representations of sentences are stated. However, none of those  approaches define precisely the actual base types or sorts to be used in the lexicon.\par
In this article, we shall discuss some of the options commonly adopted by researchers in formal lexical semantics, and defend the view that classifiers in the languages which have such pronouns are an appealing solution, both linguistically and cognitively motivated. 
\end{abstract}

\section*{Introduction}

One of the most difficult aspect of the automated processing of human language is the phenomenon of \emph{polysemy}, the ability for words to be used for different meanings in different contexts. Relatively recent studies, such as \cite{gen-lexicon}, have held the view that polysemy is a feature that enables creativity in linguistic acts, and that the meaning of words might be deduced by the application of generative mechanisms from their contexts, via processes refining semantical composition. Instead of thinking of all words denoting individual objects as sharing the same semantic types (of \emph{entities}), advanced lexical semantics could class them along \emph{lexical sorts} according to their contextual behaviour, and a process of type-checking could infer the correct meaning from any combination of predicate and object.\par
For the computational linguist, the problem of lexical semantics thus becomes twofold:
\begin{enumerate}
\item How does the semantic composition have to be modified ?
\item How should the base types, the lexical sorts, be defined ?
\end{enumerate}

The first point has been the subject of many different and related proposals, including the authors' own framework. This paper is concerned with the second part of the problem, and propose a linguistically-motivated solution.

\section{Including Lexical Considerations into Syntactical and Semantical Parsers} 

There are some wide coverage analysers that produce complete semantic analyses expressed as logical formulae, like Boxer by Johan Bos (English) and Grail by Richard Moot (spoken Dutch, written French). In both cases, the grammar, that is, a lexicon mapping each word to several semantic categories, is statistically acquired from annotated corpora. It thus has up to one hundred categories per word, hence the parser first computes the most likely sequences of categories and parse the $n$ best. 
See \cite{Bos2008STEP2,moot10grail}.

In order to compute semantic representation, both use \emph{categorial grammars} (mutlimodal or combinatory CG) and this is not a coincidence. 
Indeed, categorial grammars allow easy transformation from syntactic categories to semantic types and from syntactic analyses to semantic analyses.

Both analysers, as well as many other practical and theoretical frameworks, rely on principles of semantical composition along with the tradition of Montague Grammar, specified in \cite{montague-proper} and refined many times since.\par

Montague Grammar assumes that words have a correspondence with terms of the simply-typed $\lambda$-calculus, with applications and abstractions given by the syntactic structure of the utterance, sentence or discourse. Those terms are constructed in a type system that use two types, $\ttt$ for truth-valued formulae, and $\eee$ for entities. In that way, all single entities share the same sort, $\eee$.

Some frameworks and analysers also add the base type $\sss$, for indices of possible worlds, and the abstract sort $\event$ for events. However, linguistic entities still share the single sort $\eee$.\par

Considerations of lexical semantics provide compelling arguments for different base types. Specifically, the single sort $e$ for entities can be split in several sorts, refining the type system. Consider:
\begin{exe}
\ex \begin{xlist}
	\ex \label{bark1} The hound barked.
	\ex *   \label{bark2} The vase barked.
	\ex  ? \label{bark3} The sergeant barked.
	\end{xlist}
\end{exe}

Restrictions of selection (what, according to dictionaries, noun phrases can be object to specific verbs) dictate that (\ref{bark1}) is correct, (\ref{bark2}) is difficult to admit without a clear context, and (\ref{bark3}) is acceptable, but indicates a common metaphorical usage of \emph{bark}, implying that the person referred to has certain dog-like qualities.\par
If the distinction is made by an analyser at the stage of semantic composition, using a singular sort $\eee$ for all entities does not allow to distinguish between the syntactically similar sentences. Using different sorts for animate and inanimate entities (as commonly used in dictionary definitions) will licence (\ref{bark1}) and reject (\ref{bark2})\footnote{This does not imply that sentences such as (\ref{bark2}) should never receive any semantic analysis. There are some contexts (such as fairy tales or fantasy) that can give meaning to such sentences, and strategies to deal with those and compute a correct semantics with the same compositional analysis. In order to recognise that such a special treatment is needed, however, the system still needs to detect that the use is non-standard; it is as simple as detecting a type clash}.\par 
With additional distinctions between, in this case, dogs and humans, and a flexible typing system that detects type clashes and licence certain modification to the typing of lexical entities, the metaphorical usage of the verb in (\ref{bark3}) can be detected and identified.\par

Lexical semantics also helps with the common problem of \emph{word sense disambiguation}. A common use of words pertaining to organisations such as banks, schools, or newspapers is to represent some unnamed person that is responsible for the conduct of that organisation. Consider:
\begin{exe}
\ex \label{bank1} The bank has covered for the extra expenses.
\end{exe}
(\ref{bank1}) means that someone has taken the liberty mentioned. Distinguishing between the normal use of the word (as an organisation) and this specific use (as an agent within that organisation) is only possible if the semantic system has a mean to set them apart, and a way to accomplish this is having \emph{Organisation}s and \emph{Agent}s being two different sorts of entities in the type system.\par
\cite{gen-lexicon} and other related publications present a broad linguistic framework encompassing those issues and many others related to polysemy and the creative use of words in context. It relies on a rich lexicon with several layers of information, and a many-sorted type system that help distinguish the different sorts of entities using an ontological hierarchy founded on linguistic principles.\par
The main issue is that this rich ontological type system has not been detailed, and is very much not trivial to construct, let alone that the general composition rules are missing from the original formulation.

\subsection{Rich Types and Lexical Modifiers}

The authors have defined a system for the inclusion of lexical semantics data (see \cite{mery-ndttg}, \cite{mery-jolli}, \cite{mery-thesis} and \cite{mery-recentadvances}), and some of those results have been implemented in a semantics analyser. Instead of the single sort $\eee$, we make use of many different sorts for entities that can distinguish between different linguistic behaviours.\par
Formally, this framework uses a version of the type logic with $n$ sorts, $TY_n$, detailed in \cite{muskens-meaningandp}. Without detailing functionalities outside the scope of this contribution, those $n$ sorts are used to type the different classes of entities of the lexicon. When a type clash is detected, the analyser searches for available \emph{modifiers} provided by the logical terms that would allow the analysis to proceed, and makes a note of the lexical operations used in order to compute the actual meaning of the sentence. For instance, the following sentences refer to different facets of the entity \emph{bank} (all pertaining to the finance-related concept), identifiable by the predicates used:\par
\begin{exe}
\ex \begin{xlist}
\ex \label{bank2} The bank is closed today.
\ex \label{bank3} The bank is at the next corner.
\ex \label{bank4} The bank has gone mad.
\end{xlist}\end{exe}
(\ref{bank2}) refers to one of the most common use of the word, an \emph{Organisation}, its base type. The type system maintains inferences for commonly used modifications, a very common is to refer to a physical location where the organisation is embodied, and thus the analyser would shift the type of the term to \emph{Location} in (\ref{bank3}). In (\ref{bank4}), the predicate should apply to a person, and thus the type system would look for a way to associate a person to the organisation referred to.\par

Our framework makes use of abstraction over types (and second-order $\lambda$-calculus) in order to keep track of the lexical types involved, of constraints and modifications over those types. With hand-typed lexical entries and sorts defined over a restricted domain, this approach has been implemented and tested. However, we do not have a type system covering an entire language.

As an abridged example of the analyser, consider the sample lexical entry below:
$$ 
\begin{array}{l|l|rl} 
\mbox{Lexical item} & \mbox{Main \ $\lambda$-term} & \multicolumn{1}{l}{\mbox{Modifiers}} \\ \hline 
Birmingham & birmingham^T & Id_T:T\fl T \\ 
& & t_2:T\fl P \\ 
& & t_3:T\fl Pl\\ 
\hline 
is\_a\_huge\_place & huge\_place:Pl\fl\ttt & \\ 
\hline 
voted & voted:P\fl\ttt& \\
\end{array}
$$
where the base types are defined as follows: 

\begin{tabular}{ll}
$T$ & town \\ 
$P$ & people \\ 
$Pl$ & place \\
\end{tabular} 

The sentence:
\begin{exe}
\ex Birmingham is a huge place
\end{exe}
results in a type mismatch (the predicate is of type $Pl\fl\ttt$, argument of type $T$)
$$huge\_place^{Pl\fl\ttt}(Birmingham^T))$$ 

The lexical modifier $t_3^{T\fl Pl}$ that turns a town ($T$) into a place ($Pl$) is inserted, resulting in:

$$big\_place^{Pl\fl\ttt}(t_3^{T\fl Pl} Birmingham^T))$$

Considering:
\begin{exe}
\ex \label{copred}Birmingham is a huge place and voted (Labour).
\end{exe}
In order to parse the co-predication correctly, we use a polymorphic conjunction $\Land$.
After application and reduction, this yields the following predicate:
$$ \Lambda \xi \lambda x^\xi  \ 
\lambda f^{\xi\fl\alpha} \lambda g^{\xi\fl\beta}  
(\textrm{and}^{\ttt\fl\ttt)\fl\ttt} \ (huge\_place \ (f \ x)) (voted \ (g \  x)))$$ 

Applying the argument of type $T$ and the correct modifiers $t_2$ and $t_3$, we finally obtain:

$(\textrm{and} \ (huge\_place^{Pl\fl\ttt} \ (t_3^{T\fl Pl} \ Birmingham^T)) (voted^{Pl\fl\ttt} \ (t_2^{T\fl P} \  Birmingham^T)))$ 

\subsection{The Difference between our Proposal and related Formulations} 

There are several related proposals devoted to type-driven lexical disambiguation that share many characteristics, including works by Pustejovsky, Asher, Luo and Bekki, started in \cite{pustejovsky-metaphysics}, elaborated in \cite{pustejovsky-meaning-commonsense}, extensively developped in \cite{asher-webofwords} and subject of continuing work in \cite{luo-dot} and \cite{DBLP:conf/jsai/BekkiA12}.\par
We are indebted to the authors of this proposal and many others. However, our formulation differs from the others in a significant way.\par

\paragraph{Ontological Types and Meaning Shifts}: In \cite{asher-webofwords} and other proposals, the base types are envisioned as an ontological hierarchy that derive a language-independent system of transfers of meaning. The different possible senses associated to a word are largely dependent on conceptual relations made available by its type.\par

\paragraph{Lexical-based Transformations}: In our model, while base types distinguish between different sorts and drive the disambiguation process, the availability of transformations from a sort to another is defined at the lexical level, and depends on the language. It is thus possible to define idiosyncrasies and keep a closer rein on complex linguistic phenomena. This does not exclude to have some type-level transformations for practical purposes, specifically for the factorisation of common meaning shifts (e.g. transformations that apply to all vehicles also apply to cars).

\section{Results on a restricted Domain}
\label{itipy}
As observed by a reviewer, our model does not need a wide coverage generalist semantic 
lexicon to be tested , and we actually made some experiments for a particular question (in fulfilment of a regional project Itipy), the reconstruction of itineraries from a historical (XVII-XX century, mainly XIX)  corpus of travel stories through the Pyrenees of of 576.334 words.  See \cite{LMR2012cmlf,LMRS2012taln} for details.

For such a task the grammar ought to be a wide coverage one, including a basic compositional semantics without sorts nor any lexical information. We do have such a grammar, which has been automatically extracted from annotated corpora: it is a wide coverage multimodal categorial grammar 
that is a lexicalised grammar with an easy interface with compositional semantics \`a la Montague.\par
In the absence of manually typed semantic information, the grammar only includes an automatically constructed semantic lexicon with semantic terms  that only depict the argument structure, e.g, \emph{give} has $\lambda s^\eee \lambda o^\eee \lambda d^\eee . \underline{give(s,o,d)}$ as its semantics. The actual implementation detailed in
\cite{moot10grail,moot10semi} uses $\lambda$-DRSs of Discourse Representation Theory  \cite{KR93,musk:comb96} rather than plain $\lambda$-terms in order to handle discursive phenomena.

As the task is to provide a semantic representation the paths traversed or described by the authors,
we focused on spatial and temporal semantics. Temporal semantics is handled by operators \`a la Verkuyl, that have little to do with lexical semantics, so we shall not speak about this in the present paper. 
But the semantics of space is modelled by the very framework described in the present paper. 

As expected, the sorts or base types are easier to find for a specific domain or task. 
For space and motion verbs we obviously have two sorts, namely \emph{paths} and \emph{regions}, the later one being subdivided into \emph{villages}, \emph{mountains}, and larger areas like mountains chains. 
Paths did not need to be further divided, since by the time the stories in our corpus were written people only walked on paths (that could be called trails nowadays). Nowadays for the analysing travel stories one would possibly although consider motorways, roads, trails, etc.

The principal coercion we study in this setting for the analysis of itineraries is the phenomenon known as fictive motion \cite{talmy99fictive}. 
One can say \ma{the path descends for two hours}. In order to interpret such a sentence, 
one needs to consider someone that would follow the path, although there might be no one actually following the path, and it is often difficult to tell apart whether the narrator does follow the path or not. 
Such constructions  with verbs like \ma{descendre, entrer, serpenter,...} are quite common in our corpus as examples below show: 

\begin{exe} 
\ex 
(\ldots{}) cette route qui monte sans cesse pendant deux lieues
\trans 
(\ldots{}) this road which climbs incessantly for two miles
\ex 
(\ldots{}) où les routes de Lux et de Cauterets se séparent. Celle de Lux entre dans une gorge qui vous

mène au fond d'un précipice et traverse le gave de Pau.
\trans 
(\ldots{}) where the roads to Lux and to Pau branch off. The one to Lux enters a gorge which leads you to the bottom of a precipice and traverses the Gave de Pau.
\end{exe} 

Our syntactical and semantical parser successfully analyses such examples, by considering coercion that turn an immobile object like a road into an object of type \emph{path} that can be followed. A coercion introduced by the motion verb that allow fictive motion, e.g. \ma{descendre} (\emph{descend}), construct a formula (a DRS) that says that if an individual follows the path then he will goes down. The formula introduces such an individual, bound by an existential quantifier, and it is part of discourse analysis to find out whether it is a virtual traveller or whether the character of the travel story actually followed the path. 
\cite{MPR2011cid,MPR2011taln}

With a handwritten lexicon designed for a more precise analysis of spatial semantics, our framework worked successfully, i.e., automatically obtained the proper readings (and rejected the infelicitous ones when motion event are applied to improper spatial entities). 

\subsection{The Granularity of the Type System}
The obstacle to our framework, and other related proposals, is thus the building of the system of sorts for entities. There is no real consensus on the criteria to be followed. We chose to dismiss the claims that such an endeavour is simply impossible, that compositional semantics should stick to the single-sort Montagovian $\eee$, and that any refinements should wait a phase of pragmatics or interpretation left as an exercise to the reader, as made in very blunt terms by \cite{fodor-lepore} and more reasonably by \cite{blutner-pragma}, and refuted in \cite{pustejovsky-reply} and \cite{wilks-fallacy}. We assume that a rich lexicon with a refined type system are helpful for a number of theoretical and practical applications.\par
However, in those cases, the type system is more often than not simply assumed. James Pustejosky has described how it should behave in a number of details, in publications such as \cite{pustejovsky-type}. It has never been detailed beyond the top level and some examples; as it was outlined, the system was a hierarchical ontology comprising most concepts expressed in natural language, with at least hundreds of nodes. The other proposals range between a dozen high-level sorts (\emph{animated}, \emph{physical}, \emph{abstract}\ldots) and every common noun of every language (\cite{luo-dot}), and even every possible formula with a single free variable (as formulae are derived from types, that last definition is circular). Some others, such as \cite{cooper-codygeqlic}, propose using a record type system that does away neatly with the granularity problem, as record types are redefined dynamically\footnote{However, the inclusive definition of the records type system places it beyond classical type theory, which necessitates further adaptation in the logical framework.}; or even deliberately vague approaches, arguing that a definite answer to that question would be self-defeating.

\subsection{Practical Issues with the Controversy}

While leaving the issue open is philosophically appealing, as the possibility of a definition of an actual, single metalinguistic ontology contradicts existential principles, there is a very compelling reason to pursue the matter: providing an actual implementation of a compositional lexical semantic analysis. Partial implementations, including ours illustrated in section \ref{itipy}, exist, but without a comprehensive and well-defined type system, they are largely prototypal and rely on a few hand-written types. They do prove the viability of the analysis and the interest for word sense disambiguation, but they cannot provide a really useful analysis outside the scope of very specific domains, up to now. Large-scale generic NLP applications remain out of reach. Manual or semi-automated annotations are difficult, as they have either to be restricted to a very specific domain where it is possible to define base types comprehensively, or to be few in number and thus vague and error-prone. Choices have to be made, not in order to define the essence of lexical meaning, but simply to provide testable, falsifiable models and software that can be refined for actually useful applications.\par
This does not mean that a definite set of sorts can or should be devised once and for all, but a linguistically-motivated system, adaptable and mutable, would be an important step forward.\par

\section{Type Granularity and the Classifier Systems}

Sorts should represent the different classes of objects available to a competent speaker of the language. That two words of the same syntactic category have different sorts should mark a strong difference of semantic behaviour.\par
Our type system should be useable, with a computationally reasonable number of sorts. It should nevertheless be complex enough to model the lexical differences we are looking for.\par
In short, the set of sorts used as base types should be small in cardinality, with respect to the lexicon; large in the scope of lexical differences covered, if not complete; linguistically and cognitively motivated; adaptable, and immune to idiosyncrasy.\par
There have been many studies of some linguistic features that can prove interesting candidates for such a set, including grammatical attributes (gender, noun classes\ldots) and meta-linguistic classes proposed by \cite{goddard-meta}. We have chosen to illustrate some of the properties of the classifier systems, a class of pronominal features common to several language families including many Asian languages and every Sign language.

\subsection{The Case of the Classifier Systems}

A large class of languages employ a certain category of syntactic items known as classifiers. They are used routinely for quantificational operations, most commonly for counting individuals and measuring mass nouns. Classifiers are also widely used in Sign Language (several variations) for analog purposes.\par
Classifiers are interesting, as they are used to denote categories of physical objects or abstract concepts and approximate a linguistic classification of entities. The fact that they arise spontaneously in different and wide-reaching language families, their variety and their coverage makes them good candidates for base types.
Classifiers are often present in many Asian languages (Chinese, Japanese, Korean, Vietnamese, Malay, Burmese, Thai, Hmong, Bengali, Munda), in some Amerindian and West African languages and in all Sign Languages. They are almost absent for Indo-European languages; in English a trace of a classifier is "head" in the expression "forty heads of cattle" where one can thereafter use "head(s)" to refer to some of them. 

They are used as pronouns for a class of nouns which is both linguistically and ontologically motivated. They differ from noun classes in the sense that they are much more classifiers (200--400) than noun classes used for flexion morphology and agreement ($\leq$20). Several classifiers may be used for a single noun, depending on the relevant reading. 
Classifiers are especially developed and refined for physical objects and can often stand alone with the meaning of a generic object of their class, and some nouns do not have a classifier: in such a case the noun itself may be used as a classifier. 

The notions conveyed by classifiers differ somehow from language to language. For instance, in Chinese, classifiers can be used to count individuals, measures, both, or neither (see \cite{li-classifiers} for details), the latter case being used to denote a similarity with the referred class. They are some linguistic and cultural idiosyncrasies. However, the main features of the system are common to all languages.\par

\subsection{Classifiers in French Sign Language} 

Classifiers in sign languages (see \cite{Zwitserlood2008classifiers}) are used in the language as distinct pronouns each of them applying to cognitively related nouns, in the sense that their shape evoke their visual shape or the way these entities  are used or handled. 
There are many of them for material objects, humans beings, animals,  while ideas and abstract object are gathered into wider classes. Classifiers in sign languages are hand shapes, 
that are used to express physical properties, size, position, and also the way the classified object moves. 
Here are a few examples, from French sign language (LSF): 

\begin{center}
\begin{tabular}{rl} 
Hand shape ~&~ Classifier of ... \\ 
\hline
\
horizontal M hand shape ~&~ flat object, car, bus, train (not bike) 
\\ 
vertical M hand shape ~&~ bike, horse, fish, 
\\ 
Y handshape ~&~ plane  
\\ 
C handshape ~&~ small round or cylindrical object 
\\ 
forefinger up ~&~ person   
\\ 
fist ~&~ head of a person 
\\ 
4 hand shape ~&~ a line of people 
\\ 
three crouched fingers ~&~ small animal 
\end{tabular} 
\end{center} 

The classifier used for a given object depends on what is said about the noun / entity represented by the classifier. For instance, a line of $n$ people waiting to be served at the bakery 
may be represented by $n$ fore fingers, in case for example, these $n$ people  are  individualised and 
one wants to say they were discussing, or with the 4 hand shape of one wants to says they were waiting, they were numerous etc. 

Some linguists, such as \cite{Cuxac2000}, call them pro-forms rather then classifiers. Pro-forms are analogous to pro-nouns: they stand for the form (shape) of the object: they refer to an object via its shape or part of its shape 
i.e. they depend on the aspect that is being referred to, just like the restriction of selection in lexical semantics. Polysemic mechanisms also apply to pro-forms, as different pro-forms can be used to refer to different facets of the same lexeme: e.g., a car might be referred to using a C shape (cylinder) pro-form to indicate that it is thought of as a container, or using a M shape (flat, horizontal hand, palm down) to indicate a moving vehicle.

Classifiers of sign languages are also used to identify how many objects one speaks about.

\subsection{Classifiers in Japanese}

In Japanese, the classifiers are used as counters, in a syntactic category formally known as ``numerical auxiliaries''. They are always used in conjunction with a numeral, or a pronoun referring to a numeral:
\begin{exe}
\ex \label{nin}
\gll {Otoko no Hito} ga nan Nin imasu ka ?\\
	{Male person} SUB {how-many} {counter for people} live Q ?\\
\trans 'How many men are there ?'
\end{exe}
In (\ref{nin}), \emph{Nin} is the classifier for people. The rest of the sentence makes clear that we are referring to a specific subclass, men.\par
Japanese classifiers organise a hierarchy of sorts among the lexical entities. Children or people unfamiliar with the language can get by with a dozen common classifiers, mostly used as generic classes. Competent speakers of the language are expected to use the correct classifiers in a list comprising about a hundred entries. There are also a few hundred classifiers used only in specific situations such as restricted trades or professions, or ritualistic settings. Finally, classifiers can be generated from common nouns as a creative, non-lexical use of the word.\par
Examples of classifiers in that respect include:\par

\begin{description}
\item[Generic classifiers]~
	\begin{itemize}
	\item \emph{Tsu}: empty semantic content, used to mean any object. Commonly translated as ``thing''.
	\item \emph{Nin}: people (human).
	\item Order (\emph{Ban}), frequency (\emph{Kai}), amount of time in minutes, hours, days, etc.
	\item \emph{Hai}: measure. Used to mean ``\emph{x} units of'' anything that is a mass concept, and is presented in a container (bottles of water, bowls of rice, cups of tea, etc.)
	\end{itemize}
\item[~~Common classifiers]~
	\begin{itemize}
	\item \emph{Mai}: flat or slim objects, including paper, stamps, some articles of clothing, etc.
	\item \emph{Dai}: vehicles, machines, appliances.
	\item \emph{Ko}: small things (such as dice, keys, pins) or unspecified things (their classifier is not known to the speaker or does not exist).
	\item \emph{Hon}: long and thin objects, such as pens, bottles, but also rivers, telephone calls (if they take a long time), etc.
	\end{itemize}
\item[~~Specialised classifiers]~
	\begin{itemize}
	\item \emph{Bi}: fritter and small shrimps (for fishmongers).
	\item \emph{Koma}: frames (for comic strip editors).
	\end{itemize}
\end{description}

A complete discussion of the classifier system of Japanese or any other language falls outside the scope of this publication. What we want to illustrate is that it provides  a linguistically sound classification of entities, applicable to any entity in the language (anything that can be referred by a pronoun), and derived from cognitive categories reflected by the etymology of the individual classifiers. In some cases, the classifiers are similar to words used in language that do not have a complete classifier system, such as the English \emph{head} for units of cattle (the counter \emph{Tô} for cattle and large animals is the character denoting ``head''). In others, the metaphorical reasoning behind the lexical category is apparent (\emph{Hon}, the character for ``book'' and ``root'', is used to count long things, including objects that are physically long, rivers and coasts that have a similar shape on a map, and abstract things that take a long time such as calls, movies, tennis matches\ldots).\par
The classifier system is very obviously the result of language evolution. In each language concerned, many classifiers have a different history (linguists have argued that the classifier system in Japanese, as well as in Korean and other languages of the Asia-Pacific region, has been heavily influenced by Chinese, see \cite{tsou-contact} for details). However, the grammatical need to have a categorisation of entities in order for nouns to be countable or measurable has produced classes that share similar characteristics, suggesting that they are derived from natural observation of their salient features. In other words, even if classifiers are not commonly used in linguistics to denote anything other than numerical auxiliaries, we think they provide good candidates for a type system of the granularity we are interested in.\par
Moreover, classifiers can have a behaviour similar to lexical sorts in formal lexical semantics. Entities with the same denotation can have different classifiers if they are used in different contexts. \emph{Nin} (people) can be used to count persons in many cases, but \emph{Mei} (names) will have to be used in cases the situation calls for dignity and formality. \emph{Hai} (full container) can be used to measure countable nouns, but also boats in a dismissive way (as a populist might refer to ``a shipload of migrants''). Inapplicable classifiers can be used for metaphoric usages, puns, or obscure references to the particular etymology of a word or character. The overly obsequious humility of a character might be indicated by his use of the counter for small animals (rather than people) for himself; for other persons, this is considered a grave insult (often translated as ``I am an unworthy insect'' or ``You are a mere ant to me'').

\subsection{Classifiers as Base Types: Linguistic or Cognitive Choice ?} 

What is pleasant in the choice of classifiers as base types is that they are natural both from a cognitive and from a linguistic viewpoint. They definitely are linguistic objects, since they are part or the language, being independent morphemes (words or signs). However these morphemes represent nouns, or, more precisely, refer to the relevant aspect of the noun for a particular predicate (adjective of verb), this is the reason why several classifiers are possible for a given object. Thus they also gather objects (rather than words) that resemble each other as far as a given predicate is applied to them, and this other aspect is more cognitive than linguistic. 

Clearly, the precise classifier system depends on the language, but they obey some common general properties: it suggests that the classifier system is cognitively motivated. 
An intriguing common property is that physical entities that a speaker interact with have a very precise system of classifiers, with sub classifiers (i.e., a classifier being more specific than another), thus providing a kind of ontology in the language. 
For example, human beings and animals have classifiers, and there is a richer variety of classifiers for animals usual and closer to the human species: for instance there is a specific classifier in French sign language for small animals (rabbits, rats,\ldots). 
Although it could seem natural for sign languages, because sign language is visual and gestural that physical entities have very refined classifier systems, as signs recall the 
 visual aspects of objects and  the way we handle them, it is surprising that the Asian classifier systems are actually as rich for physical objects as the one for French Sign Language. From what we read, it seems that all classifier system do represent fairly precisely the physical objects. 

For this reason we think that the classifier system is halfway between a cognitively motivated set of sorts, and a linguistic system. It is thus a good answer to our initial practical question: what should the base types be in compositional semantics if one wishes to include some lexical semantics (e.g. to limit ambiguities) to a semantic parser.

We propose building, for use by the existing analysers for syntax and semantics, a system of sorts based on the observed classifier systems and adapted to the target languages (English, French\ldots). The common use of the classifier systems indicate that they have a reasonable granularity. The classifier systems also have some limited redundancy and specialisation, that is included in our system as lexical modifiers indicating hyponymy and hyperonymy relations between sorts.

\subsection{Integrating Base Types in our Lexicon} 

Our system requires base types in order to describe lexical sorts, that is, classes of entities that behave differently from one another as semantic units. These sorts are used to categorise nouns that refer to individuals, and form the base types of our hierarchy; predicates, action nouns, adverbs and adjectives are defined by complex or functional types built from those sorts.\par
We have seen that classifiers have many desirable qualities in the description of such classes, specifically as they apply to individuals. The cover provided is extensive, and the classification is linguistically motivated; some classifiers might have an archaic origin, or other peculiar features that makes them strongly idiosyncratic, but the strength of our system lies in the accurate representation of those idiosyncrasies, and we think classifiers provide a sound entry point for the classification necessary in our lexicon.

\section{Conclusion} 

Our type-theoretical model of lexical semantics is already implemented in analysers for syntax and semantics based on refinements of Montague Grammar and categorial grammars, and has proven useful for the study of several specific linguistic issues, using restricted, hand-typed lexica. A first system being tested uses different sorts for regions, paths and times, as well as a fictive traveller, to analyse itineraries in a specific corpus of travel stories, as illustrated in section \ref{itipy}. The devising of a complete type system for each of the target languages, and thus the definition of a wide-coverage classification of entities into sorts, is a necessity for the next step: the completion of the lexicon and its semantics.\par

The base types, and the semantics for the transformations necessary for our approach, can be obtained by those methods or a combination thereof:
\begin{enumerate}
\item by statistical means (this is, however, a very difficult issue even with a very simple type system, see \cite{zettlemoyer-learning} for a discussion);
\item by hand (this is possible for restricted domains);
\item by derivation from other linguistic data.
\end{enumerate}

For that last method, we believe that the classifier systems used in various languages present the properties we would expect from such a type system. We propose to use the classifier systems as a template for classifying sorts in the target language, and are currently designing tests in order to confirm that such categories are identified as such by speakers of the language. For those languages that do not have classifiers, we are considering the adaptation of a classifier system of a language that does. Finally, if the kind of semantic analysis we want to perform is oriented towards some sorts, it is possible to use both classifiers and specific sorts.\par

\bibliographystyle{apa-good}
\bibliography{sources}


\end{document}